\documentclass[lettersize,journal]{IEEEtran}
\usepackage{amsmath,amsfonts}
\usepackage{algorithmic}
\usepackage{algorithm}
\usepackage{array}
\usepackage[caption=false,font=normalsize,labelfont=sf,textfont=sf]{subfig}
\usepackage{textcomp}
\usepackage{stfloats}
\usepackage{url}
\usepackage{verbatim}
\usepackage{graphicx}
\usepackage{cite}
\usepackage{booktabs}
\usepackage{multirow}
\usepackage[export]{adjustbox}
\usepackage{amsthm}
\usepackage{mathrsfs}

\newtheorem{definition}{Definition}

\begin{document}

\title{Automated Privacy-Preserving Techniques via Meta-Learning}

\author{Tânia Carvalho, Nuno Moniz and Luís Antunes
        % <-this % stops a space
\thanks{Faculty of Sciences of the University of Porto. Email: tania.carvalho@fc.up.pt.}% <-this % stops a space
\thanks{N. Moniz is with the Lucy Family Institute for Data \& Society, University of Notre.}
\thanks{L. Antunes is with the Faculty of Sciences of the University of Porto and TekPrivacy.}}



\maketitle

\begin{abstract}
Sharing private data for learning tasks is pivotal for transparent and secure machine learning applications.
Many privacy-preserving techniques have been proposed for this task aiming to transform the data while ensuring the privacy of individuals. Some of these techniques have been incorporated into tools, whereas others are accessed through various online platforms. However, such tools require manual configuration, which can be complex and time-consuming. Moreover, they require substantial expertise, potentially restricting their use to those with advanced technical knowledge. 
In this paper, we propose AUTOPRIV, the first automated privacy-preservation method, that eliminates the need for any manual configuration. AUTOPRIV employs meta-learning to automate the de-identification process, facilitating the secure release of data for machine learning tasks. The main goal is to anticipate the predictive performance and privacy risk of a large set of privacy configurations. We provide a ranked list of the most promising solutions, which are likely to achieve an optimal approximation within a new domain. AUTOPRIV is highly effective as it reduces computational complexity and energy consumption considerably. 
%%limit: 200 words
\end{abstract}


\section{Introduction}~\label{sec:intro}
In light of data protection regulations, developing methods that protect sensitive data while preserving its utility has become increasingly important. According to these regulations, the release or sharing of personal data is subject to a rigorous de-identification process~\cite{wp29}, which is pivotal in mitigating the risk of unauthorised disclosure of personal information and ensuring compliance with regulatory standards.
Typically, such a process involves the application of privacy-preserving techniques (PPTs) aimed at transforming data sets to minimise privacy risks, in particular the risk of re-identification, while preserving their analytical value. Effective implementation of these techniques allows organisations to facilitate responsible data-sharing practices by ensuring both ethical and legal use of sensitive information~\cite{carvalho2022survey}.

Many tools have been developed for data de-identification, allowing graphical analysis and interactive manipulation of privacy parameters to enhance the use and understanding of PPTs. Among these tools, ARX~\cite{prasser2020flexible} and Amnesia~\cite{amnesia} are particularly notable. However, applying the algorithms in these tools typically requires expertise in the field of data privacy to correctly use the appropriate PPTs for each attribute. Thus, there has been a growing interest in synthetic data as it is perceived to reduce privacy concerns without the limitations associated with traditional PPTs~\cite{machanavajjhala2008privacy}. %Synthetic data have been used in this field to deal with such pitfalls.

Deep learning-based models have become the forefront of data de-identification strategies due to their ability to generate multiple data variants while retaining similar original data properties. This approach shows promise in addressing privacy risks by replacing sensitive values with synthetic cases without the need to inspect each attribute. Unlike traditional PPTs, which are designed to transform specific attributes or instances, deep learning-based models can generate entirely new data sets that mimic the original's statistical characteristics with much less effort.  
Several tools have been developed to facilitate data synthesis. For instance, YData~\cite{ydata_package} and Gretel~\cite{gretel_package} offer online platforms that generate synthetic data, ensuring data access remains compliant with privacy regulations. Although these tools are user-friendly and have multiple synthetic data generation models available, they have some limitations. Specifically, these tools often lack privacy evaluation mechanisms. Also, the complexity of configuring and understanding the generation models and parameters can be a barrier for users without a background in data science or privacy methodologies. Moreover, the usage costs of such tools can restrict access to high-quality data.  
On top of that, it has been shown that deep learning-based models do not fully protect data~\cite{yale2019assessing} and require significant training time to accurately capture data characteristics~\cite{bird2020reducing, carvalho2022differentially}.

In addition, ensuring the high utility of the generated data -- characterised by its usefulness and effectiveness for specific tasks or applications -- is essential to preserve the integrity and reliability of the data set.
Typically, the utility evaluation includes information loss metrics and predictive performance measures for data mining/machine learning tasks. Information loss is directly quantified in a data set while predictive performance is assessed during the training of machine learning models. In the training phase, numerous machine learning algorithms can be used in combination with multiple hyperparameter optimisation techniques, which enhance the model's ability to generalise from training data to unseen data. This broadens the range of options available for evaluating the desired solution~\cite{thornton2013auto}.
Such extensive evaluations help to fine-tune the balance between data utility and privacy, ensuring that the transformations applied do not overly reduce the usefulness of the data for its intended use.

%Achieving a balance between privacy and utility w.r.t predictive performance requires careful consideration of several factors, including an understanding of data structures, their characteristics, and regulatory requirements.However, 
Finding the optimal trade-off between these conflicting goals is challenging, amplified by the scarcity of professionals with relevant expertise in both areas. It is therefore essential to assemble a multidisciplinary team dedicated to balancing privacy and utility, which is often not possible.

We therefore face four main concerns in the realm of data de-identification: \textit{i)} specialised knowledge is required to navigate the complexities of the de-identification process, \textit{ii)} deep learning-based models demand significant computational resources and memory, \textit{iii)} the complexity of the learning tasks may compromise the predictive performance and the optimisation is a tedious process, and \textit{iv)} the search for the optimal trade-off between privacy and utility requires multiple trials, which is not resource-efficient. This makes it imperative to develop privacy preservation solutions that automate the process without requiring extensive specialisation and with a minimal resource footprint. Our goal is to achieve an optimal balance among three key factors: predictive performance,  privacy and velocity. Our approach aims to optimise each aspect, ensuring that data processing is faster, the transformed data is accurate, and privacy is protected.

To address the above problems, we explore machine learning optimisation techniques to navigate large and complex spaces to quickly and effectively find optimal solutions. Also, we propose a twin meta-learning model to automate the selection of optimal PPT. One meta-model is responsible for estimating the predictive performance across various privacy configurations while the other anticipates the privacy w.r.t to re-identification risk. 
To our knowledge, there is currently no existing solution in the related literature.

In this paper, we propose the first automated privacy preservation method, called AUTOPRIV. Designed to swiftly provide a set of optimal solutions that maximise both data privacy and predictive performance, our method sets a new standard for efficiency in the data protection field.

Our main findings are summarised as follows.
\begin{enumerate}
    % \item We propose a twin meta-learning model to automate the selection of optimal PPT;
    % \item We experimentally evaluate this approach on 18 data sets, comparing state-of-the-art competing methods for predictive performance optimisation; and,
    % \item We demonstrate notable resource efficiency by drastically reducing computational costs.
    \item Maximising predictive performance does not necessarily result in a considerable negative impact on data privacy;
    \item Bandit-based optimisation approaches are the most resource-efficient and fast; and,
    %\item The interpolation method $\epsilon$-PrivateSMOTE, is the PPT most highly recommended by AUTOPRIV.
    \item There is one particular PPT that is most highly recommended by AUTOPRIV.
\end{enumerate}

The remainder of this paper is organised as follows.
Section~\ref{sec:literature} provides a literature review on data de-identification and machine learning optimisation methods. The formalisation and description of the AUTOPRIV is presented in Section~\ref{sec:appt}. All the experimental evaluation is provided in Section~\ref{sec:experiments} including the results. Section~\ref{sec:discussion} provides a discussion on such results and conclusions are presented in Section~\ref{sec:conclusion}. 

%%%%%%%%%%%%%%%%%%%%%%%%%%%%%%%%%%%%%%%%%%%%%%%%%%%%%%%%%%%%%%%%%%%%%%%%

\section{Literature Review}~\label{sec:literature}

This paper explores two main domains: de-identification and machine learning optimisation. 
In this section, we present a comprehensive review of each area, alongside our contributions to the current state-of-the-art.

\subsection{De-identification process}

% process
The de-identification process consists of three main phases~\cite{carvalho2022survey}: %\textit{i)} attribute classification, 
\textit{i)} raw disclosure risk and data utility assessment, \textit{ii)} application of privacy-preserving techniques (PPTs) mainly guided by the disclosure risk
and attribute characteristics, 
and \textit{iii)} re-assessment of disclosure risk and data utility. %Finding the optimal balance between disclosure risk and data utility is critical. 
If the balance between these two measures is not met, further refinement of the PPTs may be necessary. This iterative process is essential for successful de-identification.
% QIs
Moreover, its effectiveness heavily depends on the assumptions made about an intruder's background knowledge, particularly relevant in the selection of quasi-identifiers (QIs), such as date of birth, gender, and occupation. When combined, these QIs can form a unique signature, heightening the risk of personal information disclosure. Therefore, it is essential to apply appropriate transformations to QIs to mitigate such risks~\cite{domingo2008survey,carvalho2022survey}.

% k-anonymity, record linkage and MI
The transformed data set is then evaluated in terms of its privacy risk, of which re-identification poses the most significant threat~\cite{wp29}. Two standard measures for re-identification risk are $k$-anonymity and record linkage. $K$-anonymity~\cite{samarati2001protecting} indicates how many $k$ occurrences occur in the data set for a given combination of QI values. An intruder can single out an individual if $k = 1$. On the other hand, record linkage (or linkability)~\cite{fellegi69} aims to measure the ability of re-identification by linking two records using similarity functions.
Apart from assessing privacy in transformed data sets, 
privacy evaluation has also been conducted during the machine learning training phase. A well-known measure is Membership Inference (MI), which aims to identify whether a given data point is included in the training set~\cite{long2017towards, houssiau2022tapas}. 

% Predictive performance
When assessing data utility in terms of predictive performance, the quality of the protected data set determines its effectiveness. The closer the evaluation results are between the original and transformed data, the more utility is preserved. Commonly used metrics for evaluating predictive performance include accuracy, precision and recall~\cite{Kent_prec_acc}.

\vspace{0.5em}
\subsubsection{Strategies for de-identification}
De-identified data can be achieved by a variety of PPTs. We analyse the most popular: traditional and synthetic-based solutions.

Traditional techniques include generalisation (recoding values into broader categories), suppression (replacing values with \textit{NaN} or special character) and noise to ensure a desired level of privacy. These can be validated using various tools, such as $k$-anonymity~\cite{samarati2001protecting} or Differential Privacy (DP)~\cite{dwork2008differential}. 
A data set is $k$-anonymous if each individual in a data set is indistinguishable from at least $k - 1$ other individuals. 
%On the other hand, 
To achieve DP, the statistical results of a data set should not be affected by the contribution of any individual.

Synthetic-based solutions have recently gained prominence as an alternative to traditional approaches~\cite{machanavajjhala2008privacy}. Examples include Generative Adversarial Networks (GANs)~\cite{figueira2022survey} such as conditional GANs and variational autoencoders (VAEs), e.g., MedGAN~\cite{choi2017generating}, TableGAN~\cite{park2018data} and PATE-GAN~\cite{jordon2018pate}. However, these methods present some challenges. For example, they need to model discrete and continuous attributes simultaneously and not solve the imbalance of categorical attributes. CTGAN~\cite{xu2019modeling} was suggested to overcome such limitations. Many other deep learning-based models have emerged, in which Figueira and Vaz~\cite{figueira2022survey} detailed survey them, also including interpolation methods. 

%Moreover, when evaluating the privacy aspects of these generative methods, significant concerns have been raised. Studies have shown that medical data generated by GANs does not model outliers well and that such models are also vulnerable to membership inference attacks~\cite{yale2019assessing}. This highlights the need for careful consideration of both the utility and privacy implications when deploying these synthetic data generation models.

%Examples include Generative Adversarial Networks (GANs)~\cite{goodfellow2014generative}, variational autoencoders (VAEs)~\cite{kingma2013auto}, CTGAN~\cite{xu2019modeling} and PATE-GAN~\cite{jordon2018pate}.

For high privacy guarantees, DP has been integrated into the synthetic data generation process. For instance, DPGAN~\cite{xie2018differentially} is a differentially private GAN model which adds Gaussian noise during the training process of the discriminator. On the other hand, DP-CGAN~\cite{torkzadehmahani2019dp} uses Rényi Differential Privacy Accountant~\cite{mironov2017renyi} to track the privacy budget. Both works by clipping the gradients of discriminator loss on real and fake data separately, summing the two sets of gradients, and adding Gaussian noise to the sum. Also, PATE-GAN can generate new data with DP guarantees~\cite{jordon2019pate}. 

Additionally, classification and regression trees have been adapted for synthetic data generation~\cite{nowok2016synthpop} in which DP has been added to the tree-based models~\cite{mahiou2022dpart}. 
Furthermore, PrivBayes~\cite{zhang2017privbayes} generates artificial data using DP through a Bayesian Network.
Nevertheless, a few approaches have emerged for releasing differentially private data sets without machine learning models. ($K$, $\epsilon$)-anonymity~\cite{holohan2017k} uses DP on a single attribute by guaranteeing a certain level of $k$-anonymity and then applying the noise to each group w.r.t QI. Most recently, $\epsilon$-PrivateSMOTE~\cite{carvalho2022differentially} was proposed combining DP with an interpolation method to obfuscate the highest-risk cases, determined by the $k$-anonymity principles.

\vspace{0.5em}
\subsubsection{Current research directions}
%Although traditional techniques using generalisation and suppression preserve the truthfulness of data, it results in a reduction of its granularity, which is likely to impact the predictive performance of machine learning models~\cite{brickell2008cost}. Furthermore, 
Traditional techniques require expertise in both privacy and utility fields, as accurate attribute classification is essential to apply appropriately parameterised transformations. For instance, attributes such as \textit{Age} can be grouped into ranges of 2, 5 or 10 years, while \textit{Address} can be categorised into councils or cities. The exploration of all possible combinations creates a large solution space, which is often impractical to explore manually. %Furthermore, the user needs specialised knowledge in data de-identification algorithms as different PPTs can be employed trough different algorithms. 
Available tools for these transformations are manually configured (e.g. ARX~\cite{prasser2020flexible}), which is a time-consuming process, making it difficult to create several protected data variants and to find the optimal trade-off between privacy and utility. Given these limitations, our work focuses on synthetic-based methods.

The popularity of synthetic data generation models in the privacy field has raised some concerns about how individuals' privacy is preserved. 
A comparison of generative methods shows that medical data generated by GANs do not model outliers well and are vulnerable to MI attacks~\cite{yale2019assessing}. Similarly, Stadler et al.~\cite{stadler2022synthetic} identified vulnerabilities to MI attacks, demonstrating that synthetic data might not maintain the utility of the original data.
The vulnerability of outliers is also demonstrated through linkage attacks~\cite{trindade2024synthetic}. Additionally, the authors show that differentially private models are more efficient at protecting extreme cases but with a higher utility degradation. This is because adding noise via DP can introduce higher levels of uncertainty, posing challenges for data analysts and potentially leading to less accurate machine learning models~\cite{carvalho2022differentially}.
A significant drawback of generative models concerns their complexity, which often translates into high computational costs in terms of memory usage and extended execution times~\cite{bird2020reducing, carvalho2022differentially}. 

Similar to traditional techniques, synthetic-based solutions are also parameterised. Finding the best protected data variant remains a challenge, as different generation models offer distinct solutions. Not only the privacy guarantees may vary across different synthetisation approaches, but also the quality of synthetic data~\cite{livieris2024evaluation}. Consequently, creating a solution space with several synthetic data variants is computationally intensive. Furthermore, evaluating the predictive performance of this solution space is particularly time-consuming. Each data variant is thoroughly evaluated across multiple machine learning models, each fine-tuned with a specific set of hyperparameters in search of optimal predictive performance. 
Although this process demands less manual configuration than traditional techniques, it still requires some expert knowledge on synthetic data generation models and machine learning models to effectively evaluate their capacity to preserve utility.

Given the iterative and resource-intensive nature of the de-identification process, our primary goal is to optimise the learning process. We then use meta-learning to automate and guide the selection of the best PPT that maximises both privacy and utility. %Therefore, we leverage the meta-learning approach to learn from past experiences and guide the selection process. 
To the best of our knowledge, this is the first approach in the literature to provide an automated solution for privacy preservation.

\subsection{Machine Learning Optimisation}

%Automated Machine Learning (AutoML) aims to automate the end-to-end process of developing machine learning models, which is crucial when evaluating the predictive performance of multiple protected data variants. The main goal is to reduce the human effort placed in building accurate predictive models. It simplifies various stages such as data pre-processing, feature engineering, model selection and hyperparameter tuning. AutoML tools use advanced algorithms and optimisation techniques to automatically search and select the best-performing model configuration for a given dataset and task. This topic has been extensively surveyed~\cite{tuggener2019automated,zoller2021benchmark,vanschoren2018meta}. %kim2022survey 
%Although AutoML aims at automating the entire machine learning pipeline, 
Our main developments focus on algorithm selection and hyperparameter tuning, the so-called CASH problem~\cite{thornton2013auto}.

Grid search is a widely popular hyperparameter tuning technique known for its simplicity and effectiveness in finding optimal configurations. 
However, as grid search explores all possible combinations, it is exhaustive and computationally expensive. In contrast, random search~\cite{bergstra2012random} randomly selects hyperparameter values for a given number of iterations. Although random search does not guarantee finding the optimal hyperparameter configuration, it usually outperforms grid search, especially when computational resources are limited.

% There are more advanced approaches to algorithm selection and hyperparameter optimisation. We divide these into two groups: optimisation-based and meta-learning methods.
%Among optimisation-based methods, 
Bayesian optimisation~\cite{Mockus1978} stands out as a sophisticated approach for algorithm selection and hyperparameter optimisation. It uses a surrogate model to approximate the objective function and iteratively proposes new hyperparameter configurations based on the model's predictions. This iterative process helps to efficiently explore the hyperparameter space and identify promising configurations making it particularly effective for optimising complex models.

A distinct and more robust approach is evolutionary algorithms~\cite{coello2007evolutionary} that are based on population search algorithms, often used for multi-objective problems. These algorithms maintain a population of hyperparameter configurations, evaluate their fitness, and use genetic operators such as crossover and mutation to generate new solutions. By prioritising better solutions to reproduction, evolutionary algorithms create an iterative process of improvement.

Two more advanced techniques, successive halving~\cite{jamieson2016non} and hyperband~\cite{li2018hyperband}, are bandit-based approaches that aim to allocate computational resources (e.g. data samples or features) while identifying optimal hyperparameter configurations efficiently.
Successive halving allocates a budget uniformly to a set of hyperparameter configurations, evaluates the performance of all configurations, rejects the worst-performing half, and interactively repeats this process until only one configuration remains. Hyperband extends successive halving in combination with the random search basis by randomly sampling several hyperparameter configurations and allocating minimal resources to each configuration. These techniques allow for rapid assessment and refinement of configurations, enhancing the efficiency of finding the best solution.
 
Recently, meta-learning algorithms have emerged as a promising strategy in this area~\cite{vanschoren2018meta,tian2022meta}. Unlike traditional machine learning algorithms, which focus on learning patterns from a specific data set or task, the major advantage of meta-learning is its ability to adapt and learn unseen tasks rapidly. 
This adaptability comes from using knowledge gained from previous tasks, allowing these algorithms to quickly generalise to new scenarios without the need for extensive retraining or fine-tuning.
The essence of the meta-learning approach lies in its strategy of learning from data set properties (meta-features) and past model evaluations to extrapolate this knowledge to diverse scenarios. Meta-features can include statistical measures, data distribution properties, or domain-specific features. 
%With a meta-model, we discard configurations that are likely to perform poorly in a particular task. 
Using a meta-model, we can efficiently eliminate configurations that are likely to underperform for a given task, streamlining the process of model selection and optimisation.

\subsection{Related Work}
Besides leveraging meta-learning for hyperparameter optimisation, this approach can also facilitate cost savings by suggesting a solution for a particular task based on results from the learning process.
Our main goal is to use meta-models, akin to stacking methods, to select the best base model with optimised hyperparameters (CASH), and then combine the predictions of the base models to enhance cost-effectiveness.

In the context of data privacy, this approach was employed by Guo et al.~\cite{guo2020privacy} to improve privacy-preserving logistic regression. The authors train base learners on different data partitions and use Logistic Regression with DP as a meta-model to combine their predictions. The meta-features are based on the relative importance of feature subsets, with less noise being added to features of higher importance. While the authors show how to ensure DP in stacking, we aim to use meta-learners to recommend the PPT that achieves a good trade-off between privacy and utility. Thus, instead of focusing only on DP, we use a wide range of PPTs, since our goal is to suggest a privacy configuration for transforming data for secure data sharing rather than applying privacy constraints in the training phase. The data transformation requires meta-features based on data properties and not model-based features. Most importantly, our strategy prioritises resource efficiency, including hyperparameter optimisation, an aspect not addressed by Guo et al~\cite{guo2020privacy}.

Furthermore, in contrast to the state-of-the-art, which typically focuses on representing privacy according to the level of DP~\cite{guo2020privacy} or MI~\cite{yale2019assessing}, we provide a privacy risk assessment w.r.t the threat model, namely linkage attacks. The key distinction in our approach is the employment of a twin meta-model strategy: one model focuses on performance prediction, while the other concentrates on data privacy prediction.

In short, our efforts are divided into two major parts: \textit{i)} optimisation, which will choose the best-performing learning model, and \textit{ii)} a combined twin meta-model that will select the most suitable PPT.

%AutoML automates repetitive and time-consuming tasks, allowing domain experts to focus on the interpretation of results and decision-making instead of getting involved in technical details.

\section{Meta-Learning for Privacy-Preservation}\label{sec:appt}

\begin{figure}[htb]
\centering
\includegraphics[width=\linewidth]{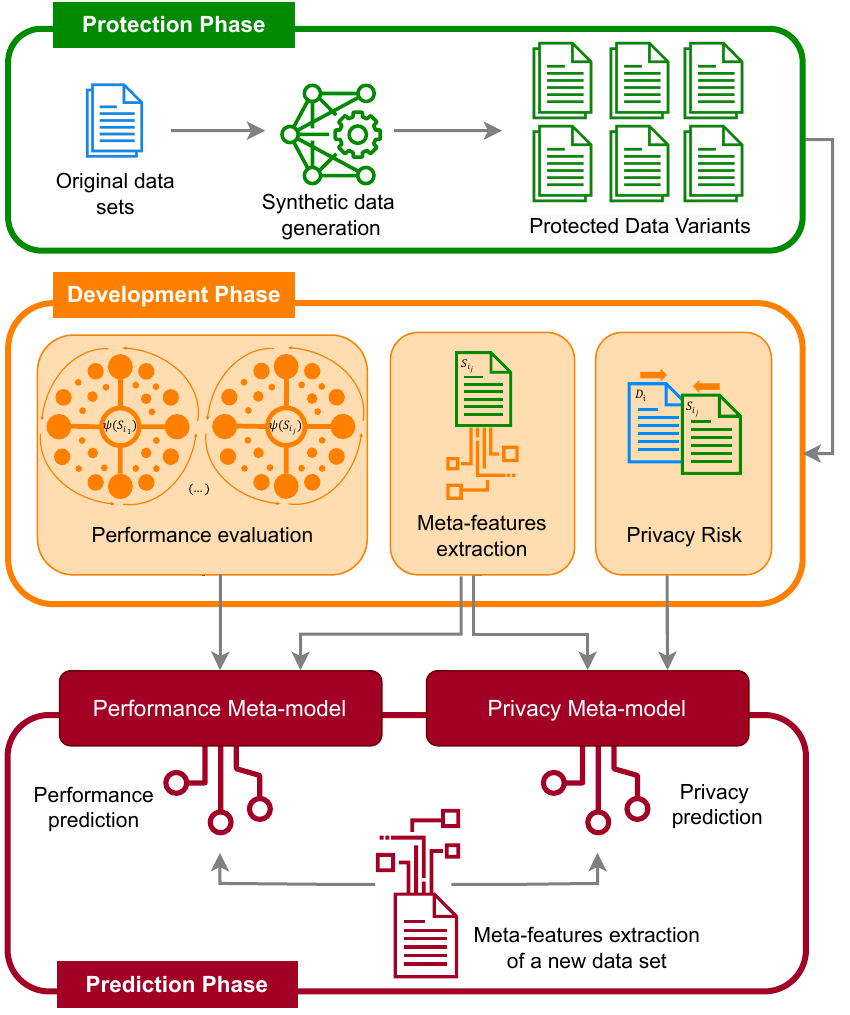}
\caption{Illustration of the AUTOPRIV method. In the protection phase, a base of original data sets $\mathscr{D}$ is transformed by a synthesis model $G$, generating multiple protected data variants. Each variant is used in the development phase to generate a twin meta-model, $\mathcal{M}$ and $\mathcal{L}$. Both models use the meta-feature description of $S_{i_j}$ ($j$ variant of the original data set $D_i$), but $\mathcal{M}$ includes the best performance within the learning configurations ($\Psi$) for all privacy configurations ($\mathcal{G}$), while $\mathcal{L}$ includes the privacy risk. In the prediction phase, we extract the meta-features of a new data set and its cross-product
of $\mathcal{G}$ to use as a predictor set for the meta-models $\mathcal{M}$ and $\mathcal{L}$.}
\label{fig:methodology}
\end{figure}

In this paper, we propose the first automated method for data de-identification. The objective is to automate the development of the solutions space to solve the task of selecting the best privacy-preserving techniques (PPTs) that offer the optimal trade-off between privacy and predictive performance. Our primary motivation is related to the costly nature of the privacy-preserving problem, where generating a set of possible solutions, especially using deep learning-based models, requires high computational resources. Secondly, the evaluation of the predictive performance involves testing all generated solutions and their respective hyperparameter optimisation. The central question is the following.
\textit{Can we leverage previous knowledge of data set characteristics and the performance/privacy of the solutions space to anticipate the performance and privacy of such solutions on new data sets?}
Figure~\ref{fig:methodology} describes the developed method which is organised in three phases: \textit{i)} protection, \textit{ii)} development and \textit{iii)} prediction phase.

\subsection{Protection Phase}
In the first phase, the objective is to generate several protected data variants using synthetic data generation approaches. While traditional synthetisation approaches change all data points, we aim to replace only the highest-risk cases, which retain a higher utility~\cite{carvalho2022differentially}.

Consider a data set $D = \{t_1, ..., t_n\}$, where $t_i$ corresponds to a tuple of attribute values for an individual's record. Let $V = \{v_1, ..., v_m\}$ be the set of $m$ attributes. A set of quasi-identifiers (QIs) consists of attribute values that could be known to the intruder for a given individual where $QIs \in V$ and is denoted as $QIs$ = $\{v_{q_{1}}, ..., v_{q_{\varphi}}\}$. For example, $QIs = \{v_{q_1}, v_{q_2}, v_{q_3}\}$ might represent an individual's gender, age, and zip code.
A tuple of selected QIs is denoted as 
$t^{QIs}_i = \{t_{i,v_{q_1}},...,t_{i,v_{q_\varphi}} : v_q \in QIs\}$, where each $t_{i,v_{q_i}}$ represents the value of a QI attribute $v_{q_i}$ in tuple $t_{i}$.
Following the $k$-anonymity basis, to ensure the protection of cases with a 50\% risk of re-identification, the protected version of $D$ ($D'$) must be at least 3-anonymous. Thus, we consider the highest-risk instances those that are single outs and cases with a 50\% risk of re-identification.
\begin{definition}[\textbf{Highest-risk selection}]
Given $D$ and $D[QIs]$, $D'$ corresponds to the highest-risk cases table, where each sequence of values in a tuple $t^{QIs}_i$ occurs at a maximum of two times in $D[QIs]$.
\end{definition}

Thus, for the $\varphi$ set of QIs, we identify the highest-risk sample in $D$, in which $T$ is the resulted data base, $T = \{D'_1, D'_2, ..., D'_\varphi\}$. Subsequently, we apply a data synthesis method denoted as $G$ to generate $\eta$ synthetic versions of each $D'$. Thus,
$G(T) = \{G_1(D'_1), G_2(D'_1),..., G_\eta(D'_1), G_1(D'_2), G_2(D'_2),..., G_\eta(D'_2),$ $..., G_1(D'_\varphi), G_2(D'_\varphi),..., G_\eta(D'_\varphi)\}$ where $G_i$ represents different synthetisation models, $i \in (1,...,\eta)$. %, producing a total of $\eta$ versions for each $D'$. 
This results in a set of $\eta \times \varphi$ synthesised data sets ($g$) representing different variants of the highest-risk samples. The combination of such models and QIs set corresponds to privacy configurations that are defined as $\mathcal{G} = \{\varrho_1,...,\varrho_g\}$. 
Finally, we add the non-highest-risk sample, to each synthetic data, $T' = \{D'_1, D'_2,..., D'_\varphi\} \cup \{D-D'_1 ,D-D'_2, ...,D-D'_\varphi\}$, where $D-D'_i$ is the non-highest-risk samples from the original data set $D$.

\subsection{Development Phase}

This phase involves three key operations: \textit{i)} develop the first stage of the meta-model, i.e., base learners to estimate predictive performance, \textit{ii)} meta-features extraction and \textit{iii)} data privacy evaluation. 

\vspace{0.5em}
\subsubsection{Predictive performance evaluation} 
In the first operation, we estimate the performance of each solution in the solution space. Consider $\mathscr{D} = \{D_1,...,D_d\}$ containing $d$ original data sets and $\mathscr{S} = \{S_{i_1},...,S_{i_j}\}$ the protected data variants for $D_i$. All base data sets are then defined as $\mathscr{C} = \{D_1,S_{1_1},...,D_2,S_{2_1},...,D_d,S_{d_g}\}$ with $d \times \eta \times \varphi$ sets ($\theta$). The respective learning task is $\mathscr{T} = \{\tau_1,...,\tau_t\}$ and the $\Psi = \{\omega_1, ..., \omega_v\}$ represent all the configurations to generate the search space to be explored in the learning process. 
Let $\mathcal{P}$ be the set of all evaluations of configurations $\omega_i$ in tasks $\tau_j$, then $\mathscr{P}_{i,j} = \mathscr{P}(\omega_i,\tau_j),i \in (1,...,v),j \in (1,...,\theta)$. The evaluation of each solution configuration is carried out using a predefined measure, e.g. AUC (Area Under the ROC Curve)~\cite{weng2008new}, and an evaluation methodology, e.g. k-fold cross-validation. The set of evaluation results for each solution configuration provides the outcome of the second operation of the development phase in the AUTOPRIV method.

\vspace{0.5em}
\subsubsection{Meta-features extraction} In this operation, we leverage the properties of $\mathscr{S}$ concerning the set of respective learning tasks $\mathscr{T}$. We use statistical functions, denoted as $\Xi_{i_j} = \xi_1,..., \xi_c$, to extract the characterisation of each protected data variant $S_{i_j}$. %Given a set of statistical functions $\Xi = \xi_1,..., \xi_c$, the information of each $S_{ij}$ corresponds to $\Xi^\Theta = \xi_1^\Theta,..., \xi_c^\Theta$. 
The information extracted is a matrix with dimensions 
$(g \times c)$, where $g$ is the number of protected data variants and $c$ is the number of statistical functions.

\begin{equation*}
\begin{bmatrix}
\xi_1(S_{1_1}) & \xi_2(S_{1_1}) & ... & \xi_c(S_{1_1}) \\
\xi_1(S_{1_2}) & \xi_2(S_{1_2}) & ... & \xi_c(S_{1_2}) \\
... & ... & ... & ... \\
\xi_1(S_{d_g}) & \xi_2(S_{d_g}) & ... & \xi_c(S_{d_g}) 
\end{bmatrix}
\end{equation*}

Based on the outcome of the two aforementioned operations, we solve the task of training a meta-model ($\mathcal{M}$) responsible for predicting the performance of all private solution configurations ($\mathcal{G}$). Our goal is to capture the connection between the meta-feature description of each protected data variant ($\Xi_{i_j}$) and the best evaluation result concerning the $\Psi$ set of learning configurations for each $S_{i_j}$. %which corresponds to the best learning configuration for $S_{i_j}$ concerning the evaluation results. %each learning configuration used,

\vspace{0.5em}
\subsubsection{Data privacy evaluation}
The last operation focuses on privacy risk evaluation through linkage ability~\cite{giomi2022unified}. The objective is to compare each protected data variant $S_{i}$ against its original data set $D_i$. 
The targeted records are a collection of $N_i$ original records randomly sampled from $D_i$.
% Suppose an intruder has some knowledge of the targets and access to $S_{a_l}$ and $D_a$, $a \in (1,...,d)$ and $l \in (1,...,g)$. % one synthetic ($\Theta$) and one obtained from public information ($\theta$), both of which contain these targets. 
The intruder's goal is to correctly match records between $S_{i_j}$ and $D_i$ given a set of QIs common to both data sets.

\begin{definition}[\textbf{Linkability~\cite{giomi2022unified}}]
For each instance in  $S_{i_j}$, an intruder finds the $k$ closest synthetic instance in $D_i$. The resulting indices are $I= (I_l,...,I_{N})$, where each $I_l$ is the set of $k$ indexes of synthetic records nearest to the $l^{th}$ target in feature subspace.
\end{definition}

If both nearest neighbour sets share the same synthetic data record in $N_{i_j}$ target, an attacker can link previously unconnected information about an individual present in $D_i$. Each successfully established link is scored as a success. The resulted outcome ($O$) is given by:
\begin{equation}
   O_{i_j}(I_l^{S{i_j}},I_l^{D_i})=\begin{cases}
      1, & \text{if $I_l^{S{i_j}} \cap I_l^{D_i}$ $\neq$ 0}\\
      0, & \text{otherwise.}
    \end{cases} 
\end{equation}

%Thus, given the $g$ set of protect data variants and $\varphi$ set of QIs, the linkability of the solutions space is $\mathcal{O} = \{O_{1}^{\Theta_1}, O_{2}^{\Theta_2},...,O_{\varphi}^{\vartheta}\}$. 
Thus, the linkability of the protected solutions space is $\mathcal{O} = \{O_{1_1}, O_{1_2},...,O_{2_1}, O_{2_2},...,O_{d_g}\}$. 
The objective is to employ a distinct meta-model ($\mathcal{L}$) to predict privacy risk. This involves utilising the previously described meta-features of each protected data variant ($\Xi_{i_j}$) to establish a correlation with the corresponding privacy risk ($O_{i_j}$). % concerning its privacy configuration ($\varrho_i$).

\subsection{Prediction Phase}
In the final phase, the goal is to recommend a private solution configuration that offers the best results in terms of predicted performance and privacy for a new data set. This phase includes the following iterations: \textit{i)} extracting meta-features from the new data set, \textit{ii)} predicting the performance for a set of  privacy configurations ($\mathcal{G}$) in the new task via meta-model $\mathcal{M}$, \textit{iii)} predicting the privacy for $\mathcal{G}$ using the meta-model $\mathcal{L}$, and \textit{iv)} generating a recommendation, i.e. an ordered list of privacy configurations that are predicted to offer the best compromise between predictive performance and privacy. Each step is described as follows.
\begin{enumerate}
    \item[i)] Given an incoming data set $D_{new}$, we use the same process in the development phase to extract its meta-features $\Xi_{new}$;
    \item[ii)] Using $\Xi_{new}$ and available private solutions configurations $\mathcal{G}$, we construct a meta-data set $\Xi_{new} \times \mathcal{G}$ that serves as the predictor set for the meta-model $\mathcal{M}$. The results concern the prediction of the performance of each privacy configuration;
    \item[iii)] The same meta-data set is also used for the meta-model $\mathcal{L}$ to predict the privacy risk, $\mathcal{O}_{new}$, across all privacy configurations;
    \item[iv)] Finally, we combine the predictions of the twin meta-model by ranking the performance and privacy risk to produce an ordered set of the averaged rank concerning the privacy configurations. The ordered set $\mathcal{S}$ is the final output of AUTOPRIV.
\end{enumerate}

\section{Experimental Evaluation}\label{sec:experiments}

This section outlines the experimental evaluation procedure. We focus on conducting a comparative analysis among multiple optimisation approaches. The objective is to identify the most effective technique and adapt it as a component of the AUTOPRIV method.
Therefore, we aim to answer the following research questions. 
\begin{enumerate}
    \item [\textbf{RQ1}] How do predictive performance and privacy risk vary among optimisation methods?
    \item [\textbf{RQ2}] Which state-of-the-art optimisation method offers the best-performing model configuration while using the fewest resources?
    \item [\textbf{RQ3}] Is there a particular PPT that dominates all of the others in terms of predictive performance?    
\end{enumerate}

\subsection{Data}
To provide an extensive experimental study, we collected 18 diverse classification data sets from the public and open data repository OpenML~\cite{OpenML2013}. These datasets span various domains, ensuring different representations of real-world scenarios. Common characteristics include:
\textit{i)} binary class target, \textit{ii)} number of instances between 1.043 and 15.545, and \textit{iii)} number of features greater than five and less than 103. The average number of instances of the retrieved data sets is 4.916. Also, there is an average of 28 numerical and two nominal attributes. For transparency and reproducibility, all datasets used in our experimental evaluation are publicly available \footnote{
https://www.kaggle.com/datasets/up201204722/3-anonymity-synthetic-data?select=original}.

\subsection{Methods}

Our experimental evaluation encompasses several methods, which we segmented into the following phases: \textit{i)} data protection, \textit{ii)} meta-features extraction, \textit{iii)} learning algorithms including optimisation approaches, and \textit{iv)} predictive performance and privacy risk evaluation. 

\vspace{0.5em}
\subsubsection{Data protection}
Regarding data transformation, we employ six synthesis methods to generate multiple protected data variants from each original data set.
We use general deep learning-based models from the \textit{SDV} library~\cite{sdv}, namely Copula GAN, TVAE, and CTGAN. % configured with the parameters: $epochs$ $\in$ $\{100, 200\}$ and $batch\_size$ $\in$ $\{50, 100\}$. 
Also, we use differentially private-based models such as DPGAN and PATE-GAN from \textit{synthcity}~\cite{synthcity} library. % with $epochs$ $\in$ $\{100, 200\}$, $batch\_size$ $\in$ $\{50, 100\}$, and $\epsilon \in \{0.1, 0.5, 1.0, 5.0\}$.
Finally, we use a different synthetisation approach, based on interpolation methods, $\epsilon$-PrivateSMOTE~\cite{carvalho2022differentially}. % with $N \in \{1, 2, 3\}$, $knn \in \{1, 3, 5\}$ and $\epsilon \in \{0.1, 0.5, 1.0, 5.0, 10.0\}$. 
Previous solutions do not include $\epsilon = 10.0$ as it fails to generate several data variants. Table~\ref{tab:synth} summarises information on the use of such synthesis methods.
%36+96+135=267
\begin{table}[!ht]
\begin{center}
    \scriptsize
    \begin{adjustbox}{max width=0.8\linewidth}

\begin{tabular}{@{}l|l@{}}
\toprule
\textbf{Synthetic Approaches} & \textbf{Parameters}                                                                                                                                                \\ \midrule
Copula GAN                    & \multirow{3}{*}{\begin{tabular}[c]{@{}l@{}}$epochs$ $\in$ $\{100, 200\}$\\ $batch\_size$ $\in$ $\{50, 100\}$\end{tabular}}                                         \\
TVAE                          &                                                                                                                                                                    \\
CTGAN                         &                                                                                                                                                                    \\ \midrule
\multirow{3}{*}{\begin{tabular}[c]{@{}l@{}}DPGAN\\ PATE-GAN\end{tabular}} & \multirow{3}{*}{\begin{tabular}[c]{@{}l@{}}$epochs$ $\in$ $\{100, 200\}$\\ $batch\_size$ $\in$ $\{50, 100\}$\\ $\epsilon \in \{0.1, 0.5, 1.0, 5.0\}$\end{tabular}} \\
                                                       
&         \\
     &    \\
\midrule
$\epsilon$-PrivateSMOTE       & \begin{tabular}[c]{@{}l@{}}$N \in \{1, 2, 3\}$\\ $knn \in \{1, 3, 5\}$\\ $\epsilon \in \{0.1, 0.5, 1.0, 5.0, 10.0\}$ \end{tabular}    \\ \bottomrule
\end{tabular}
\end{adjustbox}  
\caption{Synthetisation approaches and the respective parameter grid.}
\label{tab:synth}
\end{center}
\end{table}

All of these approaches are applied to the highest-risk cases ($k<3$) identified by the selected QIs. We assume that any attribute can be a QI given the unpredictability of the information an intruder might possess. Therefore, we collect three different sets of QIs, each consisting of a random selection of 40\% of the attributes, to simulate potential background knowledge that an intruder might have access to. We are thus considering a variety of possible threat scenarios.
Consequently, the combination of various synthetisation approaches and their parameters, targeted to the specific set of QIs, results in a solution space of 267 solutions for each original data set. This yields a total of 4806 possible privacy solutions. % across all data sets.

\vspace{0.5em}
\subsubsection{Meta-features extraction}
We use pymfe~\cite{alcobacca2020mfe} library to extract all the relevant meta-features for each protected data variant. The meta-features we analyse cover a diverse range of characteristics that are crucial for informing our meta-learning algorithms. We construct the meta-data set with 111 features, which are focused on the following groups:
\begin{itemize}
    \item General description -- includes the number of instances, attributes and classes in each data set;
    \item Statistical measures -- numerical properties that describe the data distribution;
    \item Information-theoretic -- describe discrete attributes and their relationship with the classes;
    \item Model-based -- characteristics derived from machine learning models;
    \item Landmarking -- performance of learning algorithms;
    \item Clustering -- extract information about data set based on external validation indexes;
    \item Concept -- estimate the variability of class labels among examples and the examples' density;
    \item Itemset -- correlation between binary attributes;
    \item Complexity -- estimate the difficulty in separating the data points into their expected classes; and
    \item Privacy configurations -- extracted from the data protection phase that informs privacy-preserving settings.
\end{itemize}
%We construct the meta-dataset comprising 111 features, which includes general descriptors. These include basic data set characteristics such as the number of instances and attributes; statistical metrics related to data distribution; correlations among numerical attributes; and information-theoretic measures. Additionally, model-based metrics such as performance, variability of class labels, and essential characteristics are integrated. We also incorporate the privacy configurations defined during the data protection phase.

\vspace{0.5em}
\subsubsection{Learning algorithms and optimisation}
As part of the experimental evaluation, we evaluate the predictive performance using the following methodology.
We apply four classification algorithms to evaluate the predictive performance (XGBoost, Gradient Boosting, Stochastic Gradient Descendent and Neural Network) using Scikit-learn~\cite{pedregosa2011scikit}, to test all the protected data variants. We use the early stop function 
to detect when validation set performance plateaus or declines consistently over 10 consecutive rounds.
To determine the most effective models for each algorithm, we apply a robust validation approach using a 2*5-fold cross-validation estimation of the evaluation scores. We also assess the performance of these models on test data -- 20\% of the original data set. The details of this methodology are summarised in Table~\ref{tab:algorithms}.

\begin{table}[!ht]
\begin{center}
    \scriptsize
    \begin{adjustbox}{max width=\linewidth}
\begin{tabular}{@{}l|l@{}}
\toprule
\textbf{Algorithm}  & \textbf{Parameters}                                       \\ \midrule
XGBoost and
Gradient Boosting & \begin{tabular}[c]{@{}l@{}}$n\_estimators \in \{100, 250, 500\}$\\ $max\_depth \in \{4, 7, 10\}$ \\ $learning\_rate \in \{0.01, 0.1\}$\end{tabular}                                                               \\ \midrule
Stochastic Gradient Descendent       & \begin{tabular}[c]{@{}l@{}}$alpha \in \{100, 250, 500\}$\\ $max\_iter \in \{10000, 100000\}$ \\ $eta0 \in \{0.01, 0.1\}$\end{tabular}                                                                                       \\ \midrule
Neural Network & \begin{tabular}[c]{@{}l@{}}$hidden\_layer\_sizes \in$ \{{[}$n\_feat${]}, {[}$n\_feat / 2${]}, {[}$n\_feat * 2/3${]},\\    $alpha \in \{0.01, 0.1\}$\\ $max\_iter \in \{10000, 100000\}$\end{tabular}  \\
\bottomrule
\end{tabular}
\end{adjustbox}
\end{center}    
\caption{Learning algorithms used in the experimental evaluation and respective hyperparameter grid used.}
    \label{tab:algorithms}
\end{table}

The combination of the diverse learning algorithms and hyperparameters yields a substantial configuration space.
As previously mentioned, the integration of optimisation strategies not only has the potential to save computational time and resources but also to ensure that the learning algorithms perform at their best by finding the optimal hyperparameter settings more cost-effectively.
Therefore, we use five distinct strategies for hyperparameter optimisation. Besides the traditional grid search (GS), we include random search (RS), Bayesian optimisation (BO) with 50 iterations, successive halving (SH) and hyperband (HB). Except for Bayesian Optimisation -- which we implement using Scikit-Optimize~\cite{louppe2017bayesian} -- all other strategies are executed using Scikit-learn. 
Regarding the twin meta-models, we use the Linear Regression also from Scikit-learn, to predict both performance and privacy outcomes. 

\vspace{0.5em}
\subsubsection{Predictive performance and privacy risk evaluation}

The effectiveness of predictive performance is evaluated using AUC (Area Under the ROC Curve)~\cite{weng2008new}.
We also apply statistical tests using Bayes Sign Test~\cite{Benavoli2017} to evaluate the significance of our experimental results. With such an aim, we use the percentage difference between each pair of solutions as $\frac{R_a - R_b}{R_b} * 100$ where $R_a$ is the solution under comparison and $R_b$ is the baseline solution. In the context of Bayesian analysis, ROPE (Region of Practical Equivalence)~\cite{kruschke2015bayesian} is used to specify an interval where the percentage difference is considered equivalent to the null value. In other words, ROPE relates the notion of practical equivalence as the probability of the difference of values being inside a specific interval as having practically no effect. We apply the interval [-1\%, 1\%] for ROPE. Thus, \textit{1)} if the percentage difference of the specified metric between solutions \textit{a} and \textit{b} (baseline) is greater than 1\%, solution \textit{a} outperforms \textit{b} (win); \textit{2)} if the percentage difference is within the specified range, they are of practical equivalence (draw); and \textit{3)} if the percentage difference is less than -1\%, the baseline outperforms solution \textit{a} (lose).

Concerning privacy risk we use \textit{Anonymeter} library~\cite{anonymeter} for linkability assessment. Our evaluation method involves comparing each protected data variant to the original dataset, focusing specifically on a predefined set of QIs. 
This evaluation requires a control data set, which we use as the test data set (20\% of the original data), and $k$, which indicates the nearest neighbour search space; we use $k=10$. 

\subsection{Results}
\begin{figure*}[!htb]
\centering
  \subfloat{\includegraphics[height=2in]{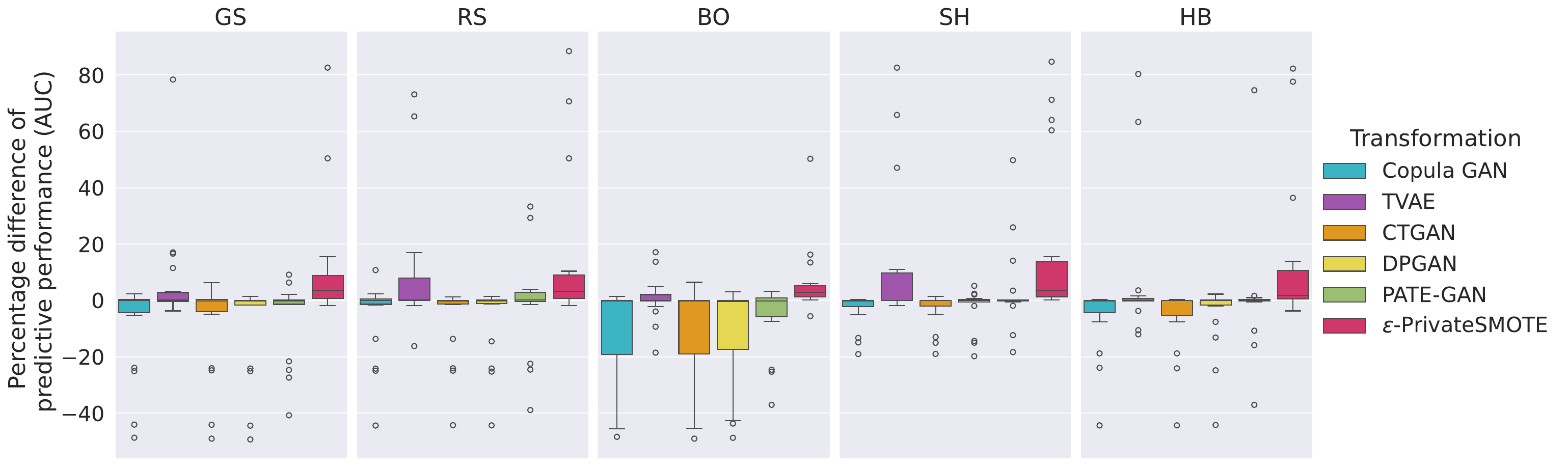}}
  \label{subfig:fig1}
\hfil
  \subfloat{\includegraphics[height=2in]
  {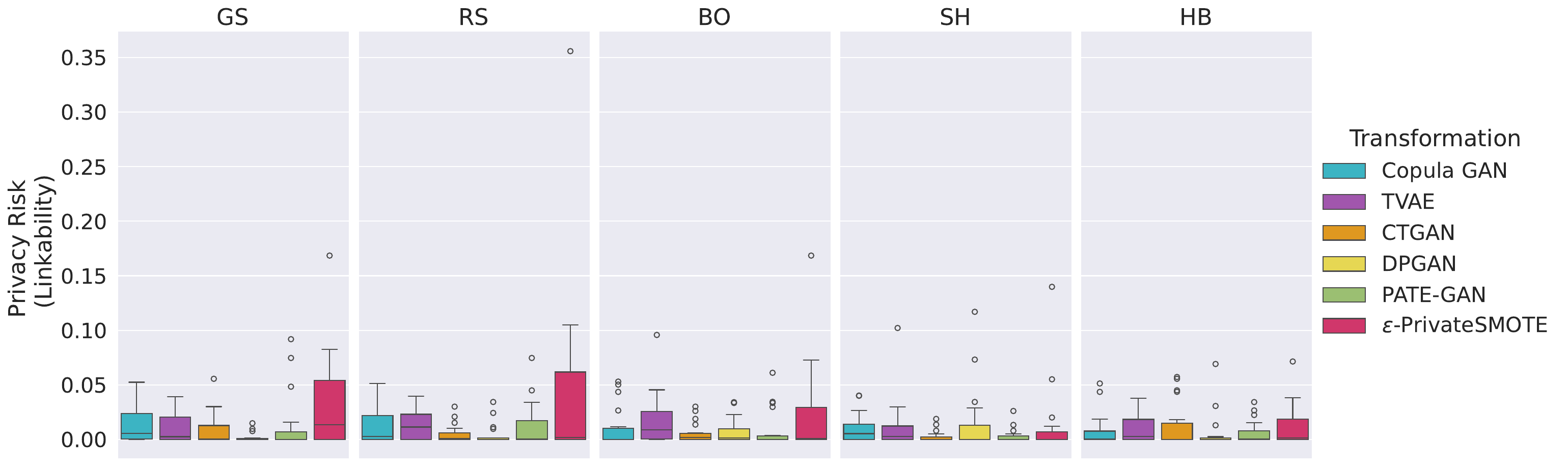}}
  \label{subfig:fig2}
\caption{Best predictive performance results (top) and corresponding privacy risk (bottom) for all hyperparameter optimisation approaches.}
\label{fig:performance_risk}
\end{figure*}

The first set of results from the experimental evaluation focuses on an overall assessment of predictive performance and privacy risk. We aim to explore the behaviour of different transformation techniques and optimisation strategies in their dual capacity of improving predictive performance and protecting privacy.
Concerning predictive performance, the results refer to the out-of-sample performance of the best models for each protected data variant, estimated by cross-validation. We then calculate the percentage difference between the best-estimated model for each protected data variant and the baseline -- best model for each original data set.
Given the diversity of solutions for each PPT, we select the best transformation in terms of predictive performance for each original data set. Figure~\ref{fig:performance_risk} illustrates such a result alongside the corresponding values for linkability for all the tested hyperparameter optimisation strategies.

Most transformation techniques appear to produce similar patterns across different hyperparameter optimisation strategies, except for Bayesian optimisation. %Although our experimental evaluation is extensive, 
The variability observed with Bayesian optimisation can be attributed to the characteristics of the data sets and the search space. Typically, this type of optimisation generally performs more effectively in lower-dimensional search spaces~\cite{seeger2004gaussian}. All the remaining appear to manage a greater control of variability, particularly successive halving, which seems to minimise extreme negative performance dips better than other strategies.

In evaluating the performance of PPTs, it is observed that Copula GAN, CTGAN, DPGAN and PATE-GAN are more susceptible to a higher negative impact on performance, which might reflect their sensitivity to specific settings. On the other hand, TAVE and $\epsilon$-PrivateSMOTE seem to exhibit more consistent performance patterns across different hyperparameter optimisation methods suggesting a greater stability in their outcomes. Additionally, these two transformation techniques show more positive performance deviations, indicating potential better utility in specific configurations. 

Despite that, $\epsilon$-PrivateSMOTE tends to exhibit less extreme fluctuations in performance, making it potentially more reliable under varying hyperparameter optimisation conditions. Also, it provides the highest positive predictive performance among all evaluated techniques.
This is mainly due to the data augmentation, as $\epsilon$-PrivateSMOTE duplicates/triplicates the highest-risk cases using an over-sampling strategy, based on the nearest neighbour algorithm.
Unlike deep learning techniques, which generate new cases within the overall distribution, $\epsilon$-PrivateSMOTE was designed to replace and replicate targeted records focusing solely on case proximity.
%Thus, $\epsilon$-PrivateSMOTE was designed to replace and replicate targeted records focusing solely on case proximity, whereas deep learning techniques generate new cases within the overall distribution.
However, since $\epsilon$-PrivateSMOTE uses a nearest neighbours algorithm, the generated cases are very similar to the original data points, which may be reflected in the linkability results. This technique presents more variability than the remaining transformation techniques with a higher deviation in grid search, random search and Bayes optimisation strategies.
We also observe that all transformation techniques generally maintain low linkability values across all optimisation strategies, suggesting effective privacy preservation with potential higher predictive performance gains for many configurations.
 
Overall, the results suggest that bandit-based approaches, particularly successive halving and hyperband, potentially provide the best trade-off between predictive performance and linkability, addressing the research question (\textbf{RQ1}) regarding how these metrics vary among optimisation methods. These strategies achieve comparably high performance with lower variability and maintain lower linkability risk.
Furthermore, the median of the linkability for all approaches is close to zero, less than 2\%, a value that is generally considered negligible in terms of privacy risk.

%However, the noise added by $\epsilon$-PrivateSMOTE results in a slightly higher privacy risk, since the linkability approach searches for the $k$-closest record, and more records may increase the linkability. Nevertheless, we highlight the uncertainty around the 

\begin{figure}[!htb]
\centering
\includegraphics[width=.9\linewidth]{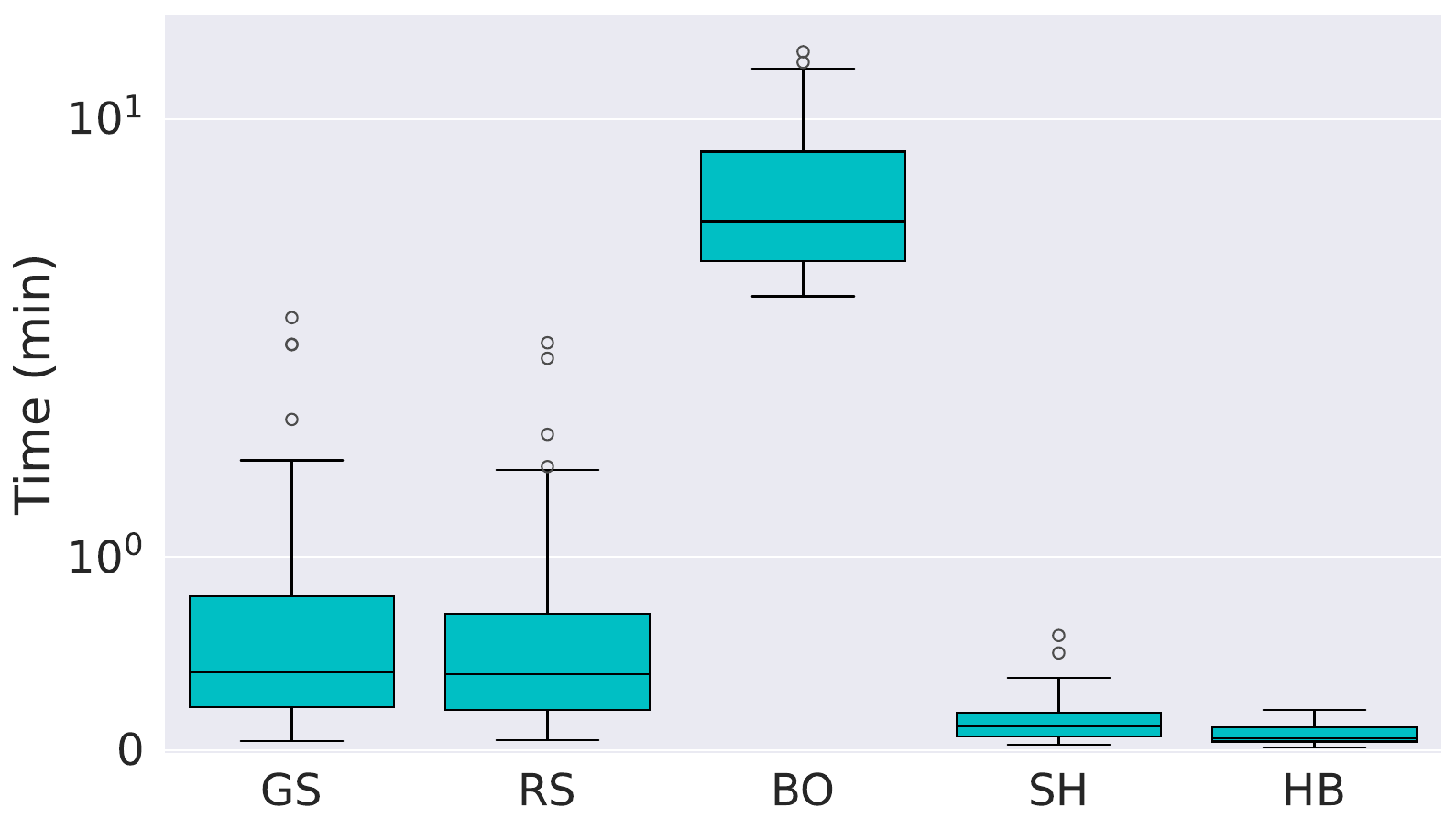}
\caption{Runtime of the hyperparameter optimisation strategies.}
\label{fig:time}
\end{figure}

However, such outcomes do not provide sufficient guidance for the selection of an optimisation strategy due to the high similarity between all strategies.
Therefore, Figure~\ref{fig:time} provides a comparison of the running time for each optimisation approach on the same set of privacy-preserving solutions to highlight differences in computational efficiency.

%An obvious result is the high runtime of Bayesian optimisation, mainly due to the iterative process, which can consume more resources to update the probabilistic model. 
The figure clearly demonstrates the high runtime associated with Bayesian optimisation, which can be attributed primarily to its iterative process that requires more resources to continually update its probabilistic model. The complexity of refining and evaluating the surrogate model at each iteration contributes to these prolonged computation times~\cite{snoek2012practical}.

On the other hand, grid search and random search show a comparable computation time between them, yet a drastically lower median time than Bayesian optimisation.
Lastly, successive halving and hyperband demonstrate the most efficient median time, indicating not only faster but also more consistent performance compared to other optimisation strategies.

\begin{figure}[!htb]
\centering
\includegraphics[width=.9\linewidth]{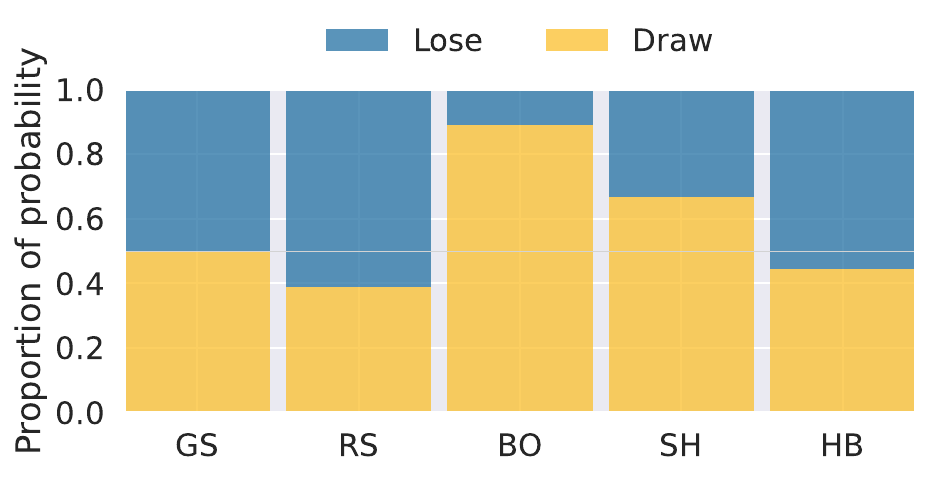}
\caption{Comparison between the competing optimisation strategies and the \textit{oracle} configuration. Shows the probability that each candidate solution significantly loses, draws or wins against the oracle according to the Bayes Sign Test for predictive performance.}
\label{fig:bayes_optype}
\end{figure}

Since we aim to reduce the computational cost of finding the optimal solution, we are interested in using one of the bandit-based approaches. To inform this decision, we present a comparison in Figure~\ref{fig:bayes_optype}, which reports the results of the Bayes Sign Test for predictive performance. This test helps to infer the statistical significance of the differences observed between paired strategies, allowing us to better understand which optimisation strategy may offer the best balance of performance and efficiency.
Thus, we establish the \textit{oracle} model and use it as a baseline. We assume that a strategy can consistently estimate the best possible model in out-of-sample performance from the explored hyperparameter configurations. Figure~\ref{fig:bayes_optype} shows the best-estimated model for each optimisation strategy and the \textit{oracle} for each data set. 
%The \textit{oracle} refers to the optimal optimisation strategy. 
In this context, the \textit{oracle} represents the hypothetical ideal optimisation strategy that consistently achieves the best possible outcomes.

%\vspace{1em}
Results show that Bayesian optimisation offers a notable advantage by achieving 90\% of practical equivalence to the best possible outcome. However, as previously noted, it demands greater computational resources. In contrast, random search shows a higher probability of losing against the \textit{oracle} with an intermediate median of runtime.
Finally, focusing on bandit-based approaches, we observe a significant advantage of successive halving over hyperband. Successive halving not only requires less computational effort but also achieves competitive predictive performance compared to the \textit{oracle}. 

Thus, in addressing \textbf{RQ2} on which optimisation method provides the best-performing model configuration with minimal resource usage, our findings highlight the potential of successive halving as a highly effective method for efficiently optimising model configurations. This efficiency makes successive halving particularly promising, as it effectively balances the three key vectors of predictive performance, privacy and velocity.
Given its demonstrated efficiency and effectiveness, we further explore the outcomes of successive halving in the context of meta-learning. Specifically, we will use them as input to the performance meta-model designed to predict the performance of new data sets. 
%Therefore, we explore the outcomes of successive halving for meta-learning, namely on the performance meta-model to predict the performance of a new data set.
%, addressing the \textbf{RQ2} on which the optimisation method delivers the best-performing model configuration with minimal resource usage. 

\begin{figure}[!htb]
\centering
\includegraphics[width=.9\linewidth]{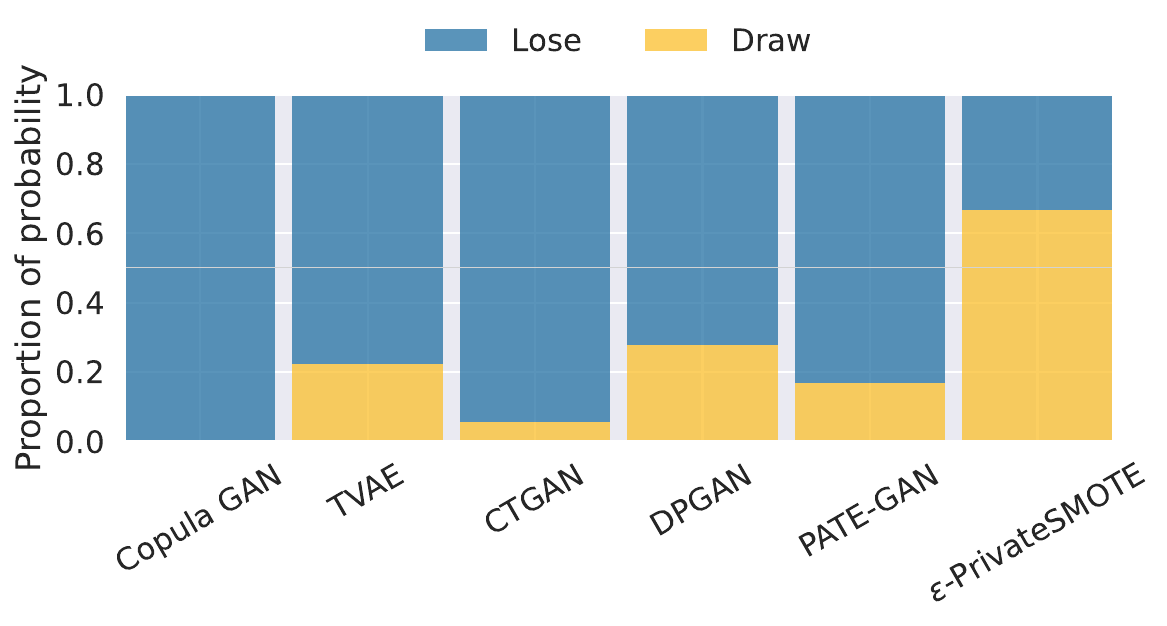}
\caption{Comparison between the best-estimated hyperparameter configuration for each transformation technique obtained through successive halving, alongside the best possible model derived from this process. Shows the probability that each candidate solution significantly draws or loses according to the Bayes Sign Test for predictive performance.}
\label{fig:halving}
\end{figure}

Regarding the dominance of a particular PPT, Figure~\ref{fig:halving} illustrates a comparative analysis. This figure contrasts the best-estimated hyperparameter configuration for each transformation technique, determined through successive halving optimisation, and compares these against the best possible model derived from the same process, focusing on their out-of-sample performance. % from the explored hyperparameter configurations. 
Note that the very low linkability values observed impact the efficacy of using a Bayes Sign Test for this analysis. Typically, this test is used to detect significant differences between paired groups; however, when the underlying values (such as linkability in this case) are extremely low, the test might not provide meaningful insights. %This is because the low variability and small range of the scores can result in insufficient evidence to statistically differentiate between the performances of various methods. 
Therefore, while the test attempts to identify distinctions in performance, the minimal linkability risk values could render these comparisons statistically inconclusive. Thus, we only focus on predictive performance differences.

Most PPTs show a high percentage of losses, indicating that they frequently underperform compared to the best possible model. Conversely, $\epsilon$-PrivateSMOTE stands out with a significant proportion of draws. 
This PPT presents a probability of practical equivalence with the best possible model using successive halving higher than 60\%, whereas other PPTs show a probability of losing higher than 70\%. This performance suggests that while $\epsilon$-PrivateSMOTE does not necessarily dominate all other techniques in every scenario, it shows potential for robust performance across different settings. Consequently, $\epsilon$-PrivateSMOTE can be strongly recommended for tasks involving privacy preservation, addressing the \textbf{RQ3}. 

All the results presented were obtained using a Linux-based system equipped with the following specifications: kernel version 5.15.0-91-generic, a 2.85GHz 24-Core AMD EPYC Processor, 256GiB of RAM, and a GeForce RTX 3090.

\section{Discussion}\label{sec:discussion}

In this section, we delve into a detailed discussion of our results, exploring their potential avenues for future research. 
%In this section, we delve into a comprehensive discussion of the results and suggest potential avenues for future research exploration. This analysis aims to further understand the implications and applications of our findings.

One of the main objectives is to recommend a solution that effectively balances both predictive performance and privacy, specifically in terms of linkability. However, it is a well-known challenge that maximising one vector often results in a loss of the other~\cite{carvalho2022survey,CARVALHO2023119785}. Despite this, our results demonstrate that it is possible to enhance predictive performance without severe privacy implications (as shown in Figure~\ref{fig:performance_risk}). We achieved this by selectively synthesising only those data points that pose the highest risk, rather than the entire data set. This targeted strategy not only ensures equivalent or even superior model performance compared to the original data but also effectively minimises privacy risks across most implemented techniques.

%Moreover, our approach successfully maintains a substantial level of data utility by selectively synthesising only those data points identified as the highest risk. Instead of synthesising the entire data set, this targeted strategy ensures that we achieve equivalent or superior model performance than original data, while concurrently minimising privacy risks across most implemented techniques.

Regardless of the PPT applied, analysing the predictive performance of the protected variant remains a complex and time-consuming task, especially when the specific learning model preferred by an end user is unknown. Thus, our analysis further extends into the efficiency of different optimisation methods in handling the dual challenges of maintaining model performance and ensuring privacy.
Given our focus on binary tasks with tabular data, we initially selected four typical models suited for such tasks. The introduction of additional models would significantly increase the complexity of the CASH process. Therefore, an optimisation method that minimises the use of computational resources becomes crucial. 
In this context, while Bayesian optimisation has shown a capacity to generalise better to unseen data  (Figure~\ref{fig:bayes_optype}), it may require a level of computational resources that may not be practical in many scenarios, as evidenced by the longer run times in Figure~\ref{fig:time}. On the other hand, bandit-based approaches offer a more resource-efficient alternative. These strategies align with our goal of enhancing model performance without sacrificing privacy or incurring computational burdens.

\begin{figure}[htb]
\centering
\includegraphics[width=.9\linewidth]{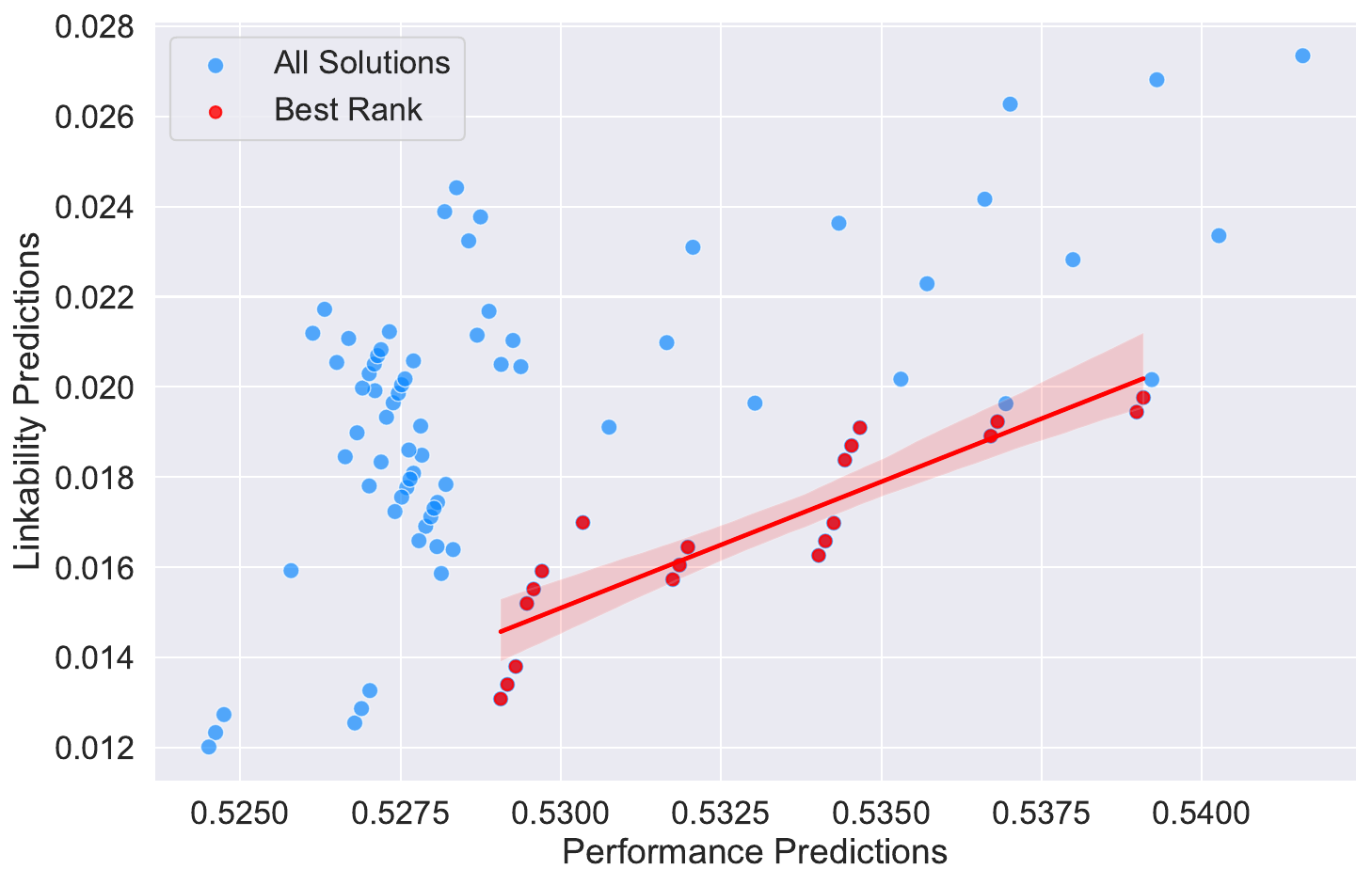}
\caption{Example of predictions of predictive performance and privacy risk (linkability), with highlighted top 20 ranked solutions.}
\label{fig:pareto}
\end{figure}

Having established the set of predictive performance and linkability risk values for all privacy configurations, we can now integrate these metrics with the meta-features of each protected variant into the twin meta-model.
The application of AUTOPRIV is illustrated in Figure~\ref{fig:pareto} where the top 20 recommendations for a new data set are highlighted in red. These recommendations represent the highest-ranking privacy solutions based on an average rank calculated between both metrics—predictive performance and linkability risk, ensuring a balance of both vectors. 

%\vspace{1em}
We focus on providing general recommendations, i.e. we predict 89 possible privacy configurations instead of the initial solution space of 267. This refinement occurs because we do not include combinations with the QIs set, since the QIs for an incoming data set are determined based on its specific characteristics and intended use.
%as we do not consider the combinations with the QI set, since the QIs for an incoming data set are defined based on data characteristics and the end use of the data depends on each case.

%We observe a tendency for the higher the predictive performance, the higher the privacy risk.
There is a positive correlation between the two metrics,  indicating that solutions with higher predicted performance tend to exhibit higher linkability. This relationship highlights the inherent trade-off between achieving high performance and maintaining privacy, even though the linkability values are close to zero. The top-ranked solutions (red dots) are those that balance this trade-off most effectively. These results demonstrate that, although a perfect balance is difficult to achieve, certain privacy configurations do offer viable compromises.
Thus, from the top 20 averaged rank recommendations, we can select a PPT whose parameterisation best aligns with the specific objectives, whether the priority is to maximise predictive performance or to prioritise enhanced privacy. 

We present in Table~\ref{tab:rank} an overview of the five highest-ranked solutions selected from a pool of 20. This table details the recommended techniques, along with their respective parameters and corresponding evaluation metrics.

\begin{table}[!ht]
\begin{center}
    \scriptsize
% \usepackage{booktabs}
\begin{adjustbox}{max width=.9\linewidth}
% \usepackage{booktabs}
% \usepackage{multirow}
\begin{tabular}{@{}l|ccc|ccc@{}}
\toprule
\textbf{Technique}            & \textit{\textbf{N}} & \textit{\textbf{knn}} & \textit{\textbf{$\epsilon$}} & \textbf{\begin{tabular}[c]{@{}c@{}}Performance\\ Predictions\end{tabular}} & \textbf{\begin{tabular}[c]{@{}c@{}}Linkability\\ Predictions\end{tabular}} & \textbf{\begin{tabular}[c]{@{}c@{}}Averaged\\ Rank\end{tabular}} \\ \midrule
\multirow{5}{*}{$\epsilon$-PrivateSMOTE} & 1                   & 3                     & 0.1                       & 0.563947                         & 0.014976                         & 71.0      \\
                              & 2                   & 3                     & 0.1                       & 0.561671                         & 0.014444                         & 71.0      \\
                              & 1                   & 3                     & 0.5                       & 0.564051                         & 0.015295                         & 69.5      \\
                              & 2                   & 3                     & 0.5                       & 0.561776                         & 0.014763                         & 69.5      \\
                              & 3                   & 3                     & 0.5                       & 0.559500                         & 0.014230                         & 69.0      \\ \bottomrule
\end{tabular}
\end{adjustbox}  
\caption{Privacy configurations of the top 5 ranked solutions.}
\label{tab:rank}
\end{center}
\end{table}

As previously demonstrated in Figure~\ref{fig:halving}, $\epsilon$-PrivateSMOTE emerges as a highly recommended option from AUTOPRIV, standing out as the best privacy-preserving solution. 

In summary, AUTOPRIV offers the following advantages for users: \textit{i)} \textbf{efficiency in data protection}, it simplifies the process of generating protected data variants, by reducing computational resources and time required for data generation; \textit{ii)} \textbf{resource optimisation}, it eliminates the massive computational burden and time requirements of analysing different PPTs w.r.t predictive performance; and, \textit{iii)} \textbf{accessible for non-experts}, it allows users with less expertise in data de-identification to easily apply the most appropriate privacy configuration to their data set, ensuring optimal privacy preservation according to specific data characteristics.

Despite the potential of AUTOPRIV, we still face the following drawbacks. 
Developing a robust recommendation system requires a considerable amount of historical data, which in turn necessitates additional development efforts and computational resources for training both the base learners and the meta-models. To overcome this, we can use meta-learning for hyperparameter optimisation~\cite{brazdil2022metalearning}. This approach leverages warm-starting, using meta-knowledge to suggest the most appropriate starting points for the optimisation search. It can guide the hyperparameter optimisation process more efficiently, reducing the number of trials needed to find optimal hyperparameter configurations. This can lead to significant savings in both computational resources and time as this strategic application of meta-learning not only optimises system performance but also addresses the scalability challenges inherent in developing advanced recommendation systems.

Additionally, the effectiveness of a meta-learning stacking strategy for cost reductions may vary depending on the domain and model configurations. This approach may not consistently generalise well or provide substantial benefits in every scenario. We will continue our efforts to research and develop advanced meta-learning techniques to more effectively tackle these challenges and enhance the adaptability and efficiency of our methodology across diverse applications.

The Python code necessary to replicate the results shown in this paper, along with detailed instructions for implementing the AUTOPRIV method on a new data set, is available at https://github.com/tmcarvalho/AUTOPRIV.

\section{Conclusion}\label{sec:conclusion}
In this paper, we propose AUTOPRIV, the first automated approach for privacy preservation based on meta-learning. Our method effectively addresses the challenges associated with the computational complexity of generating synthetic data variants and the considerable demands on resources and time required to evaluate predictive performance on a significant number of possible combinations. We compare state-of-the-art optimisation strategies to accelerate the process of evaluating predictive performance on multiple machine learning configurations.
We then combine a twin meta-model using meta-features, the best optimisation method outcomes and privacy risk assessment.
The main findings highlight the capability of AUTOPRIV to quickly predict both predictive performance and privacy criteria, thereby eliminating the need for extensive experimental trials with various privacy configurations. Furthermore, it reduces the requirement for expert knowledge in data de-identification, simplifying the process for users. 
The potential of AUTOPRIV to recommend appropriate privacy-preserving techniques for new data sets underscores its practical relevance and potential impact on real-world machine learning applications, making it a valuable tool for researchers and practitioners alike.

%\section*{Acknowledgments}
%This should be a simple paragraph before the References to thank those individuals and institutions who have supported your work on this article.

%{\appendix[Proof of the Zonklar Equations]
%Use $\backslash${\tt{appendix}} if you have a single appendix:
%Do not use $\backslash${\tt{section}} anymore after $\backslash${\tt{appendix}}, only $\backslash${\tt{section*}}.
%If you have multiple appendixes use $\backslash${\tt{appendices}} then use $\backslash${\tt{section}} to start each appendix.
%You must declare a $\backslash${\tt{section}} before using any $\backslash${\tt{subsection}} or using $\backslash${\tt{label}} ($\backslash${\tt{appendices}} by itself starts a section numbered zero.)}


%\section{References Section}
%You can use a bibliography generated by BibTeX as a .bbl file.
 %BibTeX documentation can be easily obtained at:
 %http://mirror.ctan.org/biblio/bibtex/contrib/doc/
 %The IEEEtran BibTeX style support page is:
 %http://www.michaelshell.org/tex/ieeetran/bibtex/
 
%\section{Simple References}
%You can manually copy in the resultant .bbl file and set second argument of $\backslash${\tt{begin}} to the number of references
% (used to reserve space for the reference number labels box).

\bibliographystyle{IEEEtran}
\bibliography{bibfile}

\vfill

% Biography text here without a photo.

\begin{IEEEbiography}[{\includegraphics[width=1in,height=1.25in,clip,keepaspectratio]{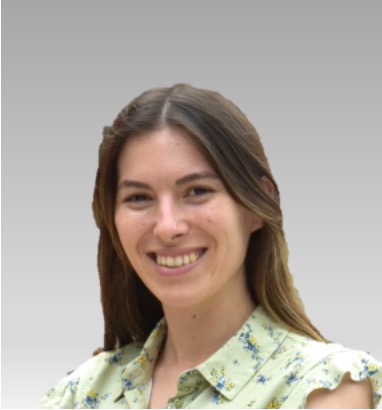}}]{Tânia Carvalho} is a Ph.D. candidate in Computer Science at Faculty of Sciences of the University of Porto. Her research focuses on the intersection of machine learning and data privacy. 
She received the EPIA23 Best Paper Award for “A Three-Way Knot: Privacy, Fairness, and Predictive Performance Dynamics" at the Portuguese Conference on Artificial Intelligence.
\end{IEEEbiography}

\vskip 0pt plus -1fil

\begin{IEEEbiography}[{\includegraphics[width=1in,height=1.25in,clip,keepaspectratio]{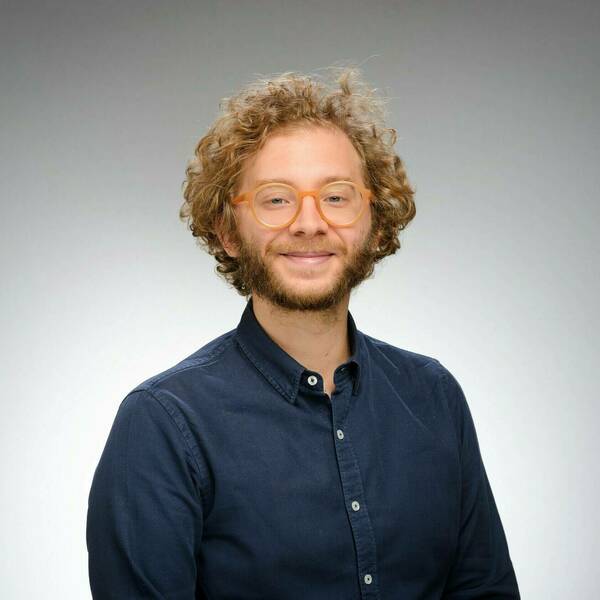}}]{Nuno Moniz}
is an Associate Research Professor at the Lucy Family Institute for Data \& Society, Director of the Notre Dame-IBM Technology Ethics Lab, and the Associate Director for the Data, Inference, Analytics, and Learning (DIAL) Lab. Moniz’s research focuses on imbalanced learning and particularly in imbalanced regression, as well as data privacy and model interpretability. He holds a PhD in Computer Science from the University of Porto. %He is particularly interested in interdisciplinary efforts to understand the real-world impact of intelligent systems.
\end{IEEEbiography}

\vskip 0pt plus -1fil

\begin{IEEEbiography}[{\includegraphics[width=1in,height=1.25in,clip,keepaspectratio]{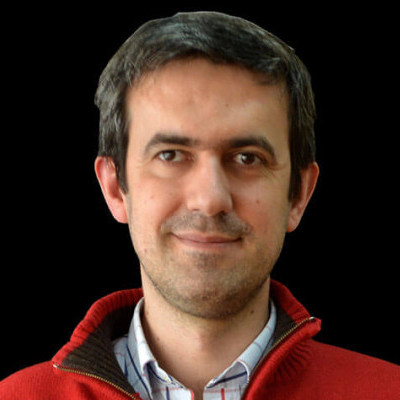}}]{Luís Antunes}
is a Full Professor at the Department of Computer Science at the Faculty of Sciences of the University of Porto. He is also the Director of the Competence Center in Cybersecurity and Privacy at the University of Porto and a member of the Board of Directors of the Association for the Progress of Business Management. Antunes develops research activity in the area of computer security, privacy and data protection. He regularly collaborates with the National Security Office, the National Data Protection Commission and the Attorney General's Office in the area of cybercrime. Founding partner of three companies HealthySystems, Adyta and TekPrivacy spin-offs of the University of Porto.
\end{IEEEbiography}

\end{document}